\documentclass[sigconf]{acmart}

\usepackage{url}            
\usepackage{booktabs}       
\usepackage{amsfonts}       
\usepackage{nicefrac}       
\usepackage{microtype}      
\usepackage{mathrsfs}
\usepackage{graphicx} 
\graphicspath{{plots/}} 

\usepackage{algorithm}
\usepackage{algpseudocode}
\usepackage{booktabs}       
\usepackage{pbox}           
\usepackage{geometry}
\usepackage[utf8]{inputenc} 
\usepackage[T1]{fontenc}    
\usepackage{hyperref}       
\usepackage{url}            
\usepackage{booktabs}       
\usepackage{amsfonts}       
\usepackage{nicefrac}       
\usepackage{microtype}      
\usepackage{bm}             
\usepackage{graphicx}       
\usepackage{array}          
\usepackage{booktabs}       
\usepackage{pbox}           
\usepackage{subcaption}     
\usepackage{enumitem}
\usepackage{calc}

\graphicspath{{plots/}} 

\AtBeginDocument{%
  \providecommand\BibTeX{{%
    \normalfont B\kern-0.5em{\scshape i\kern-0.25em b}\kern-0.8em\TeX}}}

\setcopyright{none}
\copyrightyear{2021}
\acmYear{2021}
\acmDOI{}

\acmConference[Marble-KDD 21]{Marble-KDD 21}{August 16, 2021}{Singapore}

\begin{document}

\title{A Map of Bandits for E-commerce}


\author{Yi Liu}
\authornote{Both authors contributed equally to this research.}
\email{yiam@amazon.com}
\affiliation{%
  \institution{Amazon.com}
 \city{Seattle}
 \state{WA}
 \country{United States}
}
\author{Lihong Li}
\authornotemark[1]
\email{llh@amazon.com}
\affiliation{%
  \institution{Amazon.com}
   \city{Seattle}
 \state{WA}
 \country{United States}
}

\renewcommand{\shortauthors}{}

\begin{abstract}
The rich body of Bandit literature not only offers a diverse toolbox of algorithms, but also makes it hard for a practitioner to find the right solution to solve the problem at hand. Typical textbooks on Bandits focus on designing and analyzing algorithms, and surveys on applications often present a list of individual applications. While these are valuable resources, there exists a gap in mapping applications to appropriate Bandit algorithms. In this paper, we aim to reduce this gap with a structured map of Bandits to help practitioners navigate to find relevant and practical Bandit algorithms. Instead of providing a comprehensive overview, we focus on a small number of key decision points related to reward, action, and features, which often affect how Bandit algorithms are chosen in practice.

\end{abstract}




\maketitle
\section{Introduction and Motivation}
\label {sec:intro}
Bandit is a framework for sequential decision making, where the decision maker (``agent'') sequentially chooses an action (also known as an ``arm''), potentially based on the current contextual information, and observes a reward signal. The typical goal of the agent is to learn an optimal action-selection policy to maximize some function of the observed reward signals. This problem is a special case of reinforcement learning (RL) \cite{rl_intro}, and has been a subject of extensive research in AI.

The main reason for extensive research of Bandit in the literature 
is its wide applications. The paper focuses on one of the most important domains, 
E-commerce, including online recommendation, dynamic pricing, supply chain optimization, among others \cite{bouneffouf2019survey}. For instance, Bandit has been used for online recommendation across companies such as Amazon, Google, Netflix, and Yahoo! \cite{Gangan}. Early applications were on optimizing webpage content suggestion such as news articles, advertisement, and marketing messages \cite{Li_2010, mvt}.
Nowadays, its applications have been extended to dynamic pricing~\cite{Misra}, revenue management~\cite{Ferreira}, inventory buying~\cite{Yuan},
as well as recommendation of various contents such as skills through virtual assistants~\cite{Upadhyay}.


The rich body of literature not only offers a diverse toolbox of algorithms, but also makes it hard for a practitioner to find the right solution to solve the problem at hand. A main challenge lies in the many choices when formulating a Bandit problem, and the resulting combinatorial explosion of problem space and algorithms. Typical textbooks on Bandits focus on designing and analyzing algorithms \cite{slivkins2019introduction, lattimore_szepesvari_2020}, and surveys on applications often present a list of individual applications \cite[e.g.,][]{Gangan}. While they are valuable resources, there is a gap in mapping applications to algorithms.

This paper aims to reduce this gap, by presenting a structured map for the world of Bandits. The map consists of a few key decision points, to guide practitioners to navigate in the complex world of Bandits to locate proper algorithms. While we use E-commerce as running examples, the map is useful to other applications. Furthermore, it is beyond the scope of the paper to provide a \emph{comprehensive} map. Instead, our map only focuses on a small number of factors that often affect how Bandit algorithms are chosen in practice.

The map entry is section 2, which assesses whether Bandit is the right formulation. Sections 3 and 4 describe the navigational details of the map, to help locate appropriate algorithms by inspecting several properties of rewards and actions of the application at hand. Section 5 complements the map with a discussion of topics that a practitioner often faces. Section 6 concludes the paper.

\section{Map Entry: Is Bandit the Right Formulation?}
\label {sec:entry}
\begin{figure*}[hbt!]
  \centering
  \includegraphics[width=0.75\textwidth]{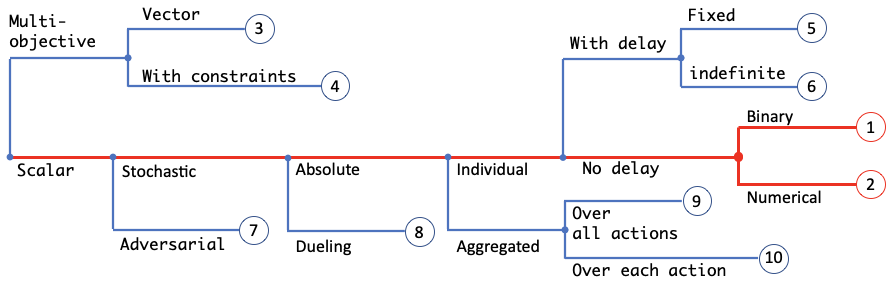}
  \vspace{-2mm}
  \caption{Bandit problems by reward properties}
  \label{fig:reward}
  \vspace{-1mm}
\end{figure*}

In typical sequential decision making modeled by Bandit, the agent repeats the following in every step: observe contextual information $x_t$, take an action $a_t$ from an action set $A_t$, and receive reward $r_t$, where $t$ denotes the step. The reward depends on $x_t$ and $a_t$. The objective of a Bandit algorithm is to recommend actions for each step to maximize the expected cumulative reward over time: $\sum_{t=1}^{T}E[r_t]$ where $T$ is the total number of steps. Suppose we want to recommend a video ($a_t$) from recently released ones ($A_t$) to customers, an application seen with Amazon Prime video, Netflix, HBO, etc. We want to consider customer features to improve the recommendation relevance: film genre (drama, romance, etc.) preference of the customer ($x_t$). The business metric of interest is total video viewers. The observed reward ($r_t$) can then be defined as $1$ if the customer watches the recommended video and $0$ otherwise. Assume stationarity and linear structure, we may model the reward as $E[r_t(x_t, a_t)] = g(w_0 + w_1 \cdot a_t + w_2 \cdot x_t \cdot a_t)$ where $w_i$'s are weights parameters and $g$ is a link function to map a linear predictor to the mean of the reward. Logit and Probit are common link functions in Bandit applications.  

The term ``Bandit'' refers to the fact that only the reward of the chosen action is observed. If rewards of \emph{all} actions are observed, the setting is called ``full-information'' \cite{cb}. For example, consider typical multi-class classification, where predicting the label for an instance can be viewed as choosing an action. It is full-information, because once we know an instance's correct label, the rewards of \emph{all} predictions are known ($1$ for the correct prediction and $0$ otherwise). One should follow a different algorithmic path of supervised learning~\cite{Elements_2009,Goodfellow_2016} if the application is in the full-information setting.

On the other hand, in the more general RL setting, the next context (``state'') may depend on previous contexts and taken actions, so the agent needs to reason with long-term rewards using more complex algorithms like Q-learning \cite{rl_intro}. For example, suppose we want to maximize sales from an online shopping website, which has search, (product) detail, add-to-cart, and checkout pages. The shopper's state on the detail page is affected by results shown earlier on the search page. 
The revenue at the end of each shopping session depends on 
information on all pages in the session. When long-term impacts are significant, Bandits may not be the best formulation, but can still be a good baseline or starting point \cite{darwin}.

\section{Navigation by Reward}
\label {sec:reward}

\begin{table*}[hbt!]
\centering
\caption{Example business applications and references for cases in Figure~\ref{fig:reward}}
\vspace{-1mm}
\label{table:reward}
\begin{tabular}{|p{0.03\linewidth} | p{0.82\linewidth} | p{0.08\linewidth} |}
\hline
No. & Business problems & References \\ 
\hline
1 & Estimate click-through rate in online ad selection; Recommend answers to questions from users to virtual assistant & \cite{BLIP, Li_2010} \\ 
\hline
2 & Recommend products on webpage to maximize revenue; Optimize inventory buying to  maximize cash flow  & \cite{riquelme2018deep, agrawal2014thompson} \\ 
\hline
3 & Recommend video content to maximize streaming customers and also paid service subscriber size  & \cite{Drugan_2013} \\ 
\hline
4 & Maximize advertisement click-through rate with cost budget for bidding; Inventory selection to maximize revenue considering inventory capacity  &  \cite{Badanidiyuru_2013, Ding_2013} \\ 
\hline
5 & N-day free-trial marketing  (Robinhood Gold, Netflix Subscription, Spotify Premium) for acquiring paid member. We do not know if users will become paid members or not until day N  & \cite{joulani2013online} \\ 
\hline
6 & Users may not watch a recommended video until later; Users may not buy a product advertised until later. Delay in a potential positive action is not time bounded. & \cite{vernade2020linear, Chapelle_2011} \\ 
\hline
7 & Malicious activities (fake reviews, click fraud, etc.) modeling in recommender system; adaptive shortest path routing  & \cite{Yang_2020, beygelzimer2011contextual,Kale_2010}\\ 
\hline
8 & Interleaved search evaluation, with relative feedback for two search results extracted from user clicks  & \cite{Yue_duel_2011}\\ 
\hline
9 & Recommendation of a combination of contents where reward cannot be easily attributed to individual content & \cite{mvt} \\
\hline
10 & Page layout selection where reward can be traced back to the component on the page & \cite{Wen15} \\
\hline
\end{tabular}
\end{table*}

The nature of reward signals in an application plays a major role in deciding the right Bandit algorithms. Figure 1 identifies six key properties of rewards that lead to 10 typical use cases in practice. The highlighted two paths are perhaps the most common. The key reward properties are: \vspace{-1mm}
\begin{description}[leftmargin=0cm]
  \item [Dimension] The reward can be one-dimension (scalar) or multi-dimension (vector). In the latter case, the task can be optimizing the vector reward (node 3), or maximizing one reward dimension subject to constraints on remaining dimensions (node 4). 
  \item [Distributional assumption] In stochastic Bandits, the reward is drawn from an unknown distribution. When the reward distribution may change slowly over time, as in many real-world applications, one can still treat it as a stochastic Bandit by constantly retraining the policy with new data. In adversarial Bandits, there is no probabilistic assumption on the reward (node 7).
  \item [Relativity] While a reward usually measures how good an action is, in some problems like ranking one can also work with relative rewards that compare actions, as in dueling Bandits (node 8).
  \item [Granularity]
When actions are combinatorial (see section~\ref{sec:action}), the reward can be for the entire action (node 9), or can provide signal for subactions that comprise the action (node 10).  The latter setting is also known as semi-bandits.
  \item [Delay] Practical limitations like software constraints may prevent rewards from being observed before the next actions have to be taken. There are two types of reward delays: bounded (node 5) and indefinite (node 6), depending on whether we have a reasonably small upper bound for the delay.
  \item [Value type] Reward can be binary (node 1) or numerical (node 2). These two leaf nodes are unique as value type must be defined for the other seven nodes to complete the reward formulation. We take this into consideration for algorithm recommendation in Table \ref{table:reward}.
\end{description}

We design the structure in Figure \ref{fig:reward} so that it covers common use cases in E-commerce. Leaf nodes 3--10 are not exhaustive as the splits are not mutually exclusive. For instance, adversarial Bandits can also be a dueling one \cite{Yue} and there can be delay in reward \cite{joulani2013online}.
In practice, however, such combinations appear uncommon.

In Table \ref{table:reward}, for each leaf note we list example business problems with suggested algorithms. Given this paper’s focus, we recommend algorithms that are empirically validated, especially those that find wide applications in practice. For leaf nodes 3--10, we may have two suggested benchmark papers when binary and numerical reward are both common. 

To find Bandit solutions for a given business problem, one can use Figure \ref{fig:reward} as a guide to land in the most relevant node, then refer to Table \ref{table:reward} for similar applications and algorithmic suggestions. We emphasize that there are no universally best algorithms, but expect the suggested references offer a good starting point for algorithm development and experimentation.


\section{Navigation by Action}
\label {sec:action}
\begin{figure}[hbt!]
  \centering
  \includegraphics[width=0.3\textwidth]{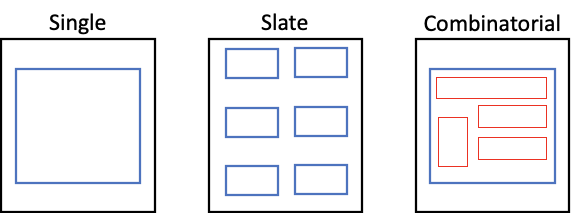}
  \vspace{-1mm}
  \caption{Illustration of three action types}
  \label{fig:actiontype}
  \vspace{-1mm}
\end{figure}

\begin{table*}[hbt!]
\centering
\caption{Example business applications and references for Single, Slate and Combinatorial Bandits}
\label{table:actiontype}
\vspace{-2mm}
\begin{tabular}{|p{0.1\linewidth} | p{0.6\linewidth} | p{0.18\linewidth}|}
\hline
Action type & Business problems & \ References \\
\hline
Single & Estimate click-through rate in online ad selection; recommend products on webpage to maximize revenue  & \cite{zhou2020neural, agrawal2014thompson, Chapelle_2011} \\ 
\hline
Slate & Return a ranked result list for user's search query; show multiple Fashion products on the landing page & Position effect: \cite{ermis2020learning, Wang_2017}; Diversity effect: \cite{Qin_2014, Yue_submodular_2011} \\ 
\hline
Combinatorial & Select components (image, title, action button, etc.) to form optimal marketing messages; whole page optimization  & \cite{mvt} \\ 
\hline
\end{tabular}
\end{table*}

\begin{figure*} [hbt!]
  \centering
  \includegraphics[width= 0.8 \textwidth]{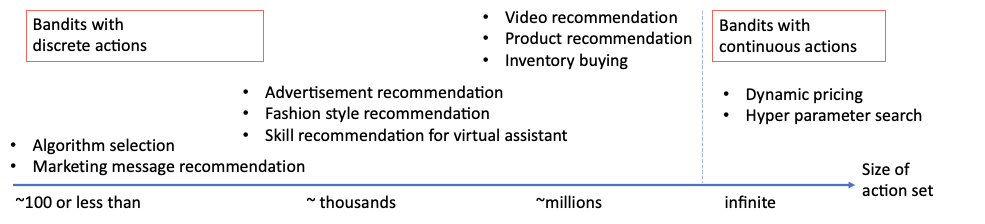}
  \vspace{-1mm}
  \caption{Example business applications by action set sizes}
  \label{fig:actionsize}
  \vspace{-1mm}
\end{figure*}

The action set $A$ also plays an important role in determining the right bandit algorithm. Here, we identify two properties: action type and action set size. The cases based on these properties are not exclusive to those identified in the previous section, but are orthogonal in many scenarios.

Figure \ref{fig:actiontype} shows three common action types: single, slate and combinatorial. In the first, the action set is a set of items. The set is often small and finite, but can also be large or even infinite. In the second, the action is a slate consisting of a ranked list of items chosen from a pool, such as ranked results for a given search query. The challenge is the exponentially many possible permutations that require more efficient algorithms. Furthermore, we need to consider two effects: position bias for actions shown at different slots and item diversity in the overall slate. In the third, the action is a combinatorial object (such as content layout on a web page), consisting of sub-actions coming from different sets. The algorithmic challenge is often in dealing with combinatorial explosions of actions, and with interaction effects between sub-actions. In the literature, slate bandits are sometimes referred to as combinatorial. Table \ref{table:actiontype} lists business problems and recommended work by these two types.


Another action property is the size of the action set. In Figure \ref{fig:actionsize}, we list representative business applications with increasing size. Many Bandit use cases have discrete actions. When the action space is small, actions ID can be considered as a categorical value and encoded as a set of binary variables in the reward formulation. When the action space becomes large, we use features to represent actions. Such action featurization not only reduces the dimension of variables but also enables generalization across actions which mitigates action cold-start problem. While action and contextual features contain different meanings, they can be handled in a similar manner. More discussions are found in section~\ref{sec:others}.
It is also common to see continuous actions in practice.  In this case, the continuous action set often has a natural distance metric, so that one can use tools like Gaussian process to solve the Bandit problem~\cite{Srinivas_2012,krishnamurthy2020contextual}.


\section{Other Topics}
\label {sec:others}
This section discusses feature engineering, offline policy evaluation, and best-arm identification, three important topics in Bandit applications to complement the previous sections.

\subsection{Feature Engineering}
In many applications, we use features to deal with large context or action sets more efficiently.
The expected reward $r$ can be written as a function of action features $\boldsymbol{{\phi}_a}$ and contextual features $\boldsymbol{{\phi}_x}$: $E[r] = f(\boldsymbol{{\phi}_a}, \boldsymbol{{\phi}_x})$.
Feature engineering in Bandit involves selection/pre-processing of $\boldsymbol{{\phi}_a}$ and $\boldsymbol{{\phi}_x}$ and their interaction terms. Linear Bandits are the most studied in the literature where $E[r]$ is assumed to be linear in the features. To model non-linearity especially when the size(s) of $\boldsymbol{{\phi}_a}$ or/and $\boldsymbol{{\phi}_x}$ is large, we can learn lower-dimension embeddings from the raw features ($\boldsymbol{{\phi}_a}$ or/and $\boldsymbol{{\phi}_x}$) and put the embeddings in the reward function instead \cite{Lin_adaptive_2020, zahavy2019deep}. Embedding generation techniques for supervised learning generally apply to Bandits~\cite{He_tree_2014}. Another way to relax the linear-reward assumption is to use non-linear Bandits where reward function becomes non-linear in feature vectors~\cite[e.g.,][]{zhou2020neural}. 

\subsection{Offline Policy Evaluation}
Testing a Bandit policy in real user traffic is often expensive, and poses risks on user experiences. It is common to evaluate a new policy offline before deploying it. A key challenge in offline policy evaluation is that we do not know how users would have reacted to actions different from the one in the log data, since the data only have rewards for selected actions. This \emph{counterfactual} nature makes Bandit offline evaluation similar to causal inference where we want to infer the average reward $E_{\pi}[r]$ (the causal effect) if policy $\pi$ is used to choose actions.
There exist effective approaches to evaluating a stationary policy, including simulation~\cite{Genie19}, inverse propensity scoring \cite{Bottou_counterfactual_2013, Strehl_loggeddata_2010}, doubly robust evaluation \cite{dudik2011doubly}, and self-normalized inverse propensity estimators \cite{Swaminathan_snip_2015}. 
Typically, offline evaluation becomes more challenging with a larger action set.
For a slate Bandit, pseudoinverse estimator is available to account for position bias \cite{Swaminathan_slate_2017}. Offline evaluation for non-stationary Bandits remains challenging \cite{Li_offline_2011,dudik2012sampleefficient}, with opportunities for further research.

\subsection{Best-arm Identification}
In some bandit applications, our goal is not to maximize reward during an experiment, but to identify the best action (e.g., best marketing campaign strategy) at the end of the experiment. This problem, known as best-arm identification \cite{kazerouni2019best, soare2014bestarm, Bubeck_pure_2010}, shares the same goal as conventional A/B/N testing~\cite{Kohavi20}, but can be statistically more efficient by adaptively selecting actions during an experiment.

\section{Conclusions}
\label {sec:conlcusions}
We presented a structured map for the world of Bandits, with the hope to guide practitioners to navigate to practical Bandit algorithms. The work is not attempted to provide a comprehensive map. Instead, it focuses on a few key decision points that often affect how Bandit algorithms are chosen. We hope it reduces the gap in connecting applications to appropriate algorithms.

\bibliography{banditmap}
\bibliographystyle{ieeetr}

\end{document}